\documentclass{article}

\usepackage{amssymb}
\usepackage{tikz}
\usepackage{appendix}
\usepackage{geometry}
\usepackage{graphicx} % Required for inserting images
\usepackage{booktabs}
\usepackage{fancyhdr}
\usepackage{hyperref}

\title{Automated Text Scoring in the Age of Generative AI for the GPU-poor}
\author{Christopher Ormerod and Alexander Kwako}
\date{christopher.ormerod@gmail.com\\
alexander.kwako@cambiumassessment.com}

\begin{document}

\maketitle

\begin{abstract}
Current research on generative language models (GLMs) for automated text scoring (ATS) has focused almost exclusively on querying proprietary models via Application Programming Interfaces (APIs). Yet such practices raise issues around transparency and security, and these methods offer little in the way of efficiency or customizability. With the recent proliferation of smaller, open-source models, there is the option to explore GLMs with computers equipped with modest, consumer-grade hardware--—that is, for the “GPU poor.” In this study, we analyze the performance and efficiency of open-source, small-scale GLMs for ATS. Results show that GLMs can be fine-tuned to achieve adequate, though not state-of-the-art, performance. In addition to ATS, we take small steps towards analyzing models' capacity for generating feedback by prompting GLMs to explain their scores. Model-generated feedback shows promise, but requires more rigorous evaluation focused on targeted use cases.
\end{abstract}

\section{Introduction}

Generative language models (GLMs), such as GPT-4 \cite{openai_gpt-4_2023} and Claude \cite{anthropic2024claude}, have demonstrated powerful performance across a variety of language and reasoning tasks. In the field of education, researchers are exploring the extent to which these models can perform tasks such as automated essay scoring \cite{xiao_automation_2024}, providing feedback to students \cite{carlson2023utilizing}, individual tutoring \cite{chevalier_language_2024}, and more \cite{gimpel_unlocking_2023}. 

Although GLMs show promise in automating certain educative tasks, there are critical limitations that hinder the possibility of wider implementation. For instance, researchers have shown that GLMs can be ``jail-broken" to bypass safety guardrails \cite{xie_defending_2023} and can disclose personally identifiable information. Large GLMs are extremely large, requiring millions of dollars to train and deploy; as such, they are highly inefficient for specialized tasks \cite{khowaja2024chatgpt}. These models are constantly being updated, sometimes leading to degraded performance \cite{chen_how_2023}, and they are only accessible via Application Programming Interfaces (APIs), which lead to issues around replicability and leave little room to conduct rigorous research. 

It is for these reasons that we shift the focus away from large, proprietary GLMs toward smaller, open-source GLMs. In this study, we focus on two educational applications: Automated Text Scoring (ATS) and providing feedback---specifically, feedback that justifies scores based on the scoring rubric. Our study is the first to demonstrate that it is possible to efficiently fine-tune such GLMs to yield high-quality scores, and that (at 
least some) feedback from fine-tuned models can explain these scores. Our data is drawn from the publicly available Automated Student Assessment Prize (ASAP),\footnote{ASAP Automated Essay Scoring: \hyperlink{ASAP Automated Essay Scoring: https://www.kaggle.com/c/asap-aes}{https://www.kaggle.com/c/asap-aes}; ASAP Automated Short Answer Scoring: \hyperlink{ASAP Automated Short Answer Scoring: https://www.kaggle.com/c/asap-sas}{https://www.kaggle.com/c/asap-sas}} which allows us to compare more easily our results to other approaches, and share our findings more broadly. More specifically, our research goals are as follows:

\begin{enumerate}
    \item Fine-tune four recently-released, relatively small (8 GB or less) open-source GLMs for Automated Essay Scoring (AES) and Automated Short Answer Scoring (ASAS).
    \item Compare the performance of these GLMs for AES and ASAS, relative to current state-of-the-art (SOTA) benchmarks. 
    \item Prompt GLMs to explain the scores that they provided based on item-specific rubrics, and characterize patterns of feedback via qualitative analysis.
\end{enumerate}

The organization of this paper is as follows: In Section \ref{sec:background}, we review the theoretical and empirical context surrounding ATS, feedback, GLM architectures, and GLM training. In Section \ref{sec:method}, we detail the characteristics of the data, models, prompts, and training methods used in this. We review results in Section \ref{sec:results}, which is divided into (A) automated scoring and (B) feedback (of essays and short answers, respectively). Finally, we discuss some of the ramifications of our findings in Section \ref{sec:discussion}, and suggest avenues for future research. In addition to this paper, for greater transparency, we make publicly available the scores and feedback generated by our fine-tuned GLMs. 

\section{Background}
\label{sec:background}

\subsection{Automated text scoring}

% Evaluating student-constructed responses is a critical component of educational assessment. Manually grading essays and short-form responses can be an arduous, time-consuming, and soul-sucking task. Automatic Text Scoring (ATS) is an application of artificial intelligence (AI) that allows teachers to spend more time teaching, preparing materials, and grading less. 
AES and ASAS have been active areas of research and development since as early as 1966 \cite{page_project_2003}. There is widespread acceptance that, when carefully constructed and monitored, AES and ASAS can deliver reliable scores \cite{lottridge_psychometric_2023}. For this reason, ATS has become common in educational assessment. 

% Within the context of student assessment, there are two main classes of ATS: Automatic Essay Scoring (AES) and Automatic Short Answer Scoring (ASAS). 
From a machine-learning perspective, both AES and ASAS are text classification problems, but from a measurement perspective, they assess different abilities and may require different approaches. For instance, rubrics for essay scoring are often designed to evaluate attributes such as organization, argumentation, grammar, and spelling in lengthier written responses. In contrast, rubrics for short answer questions focus on assessing specific knowledge and comprehension, often independent of grammatical and spelling considerations. For this reason, an approach that works well for AES may not always be suitable for ASAS and vice versa. 

There have been a plethora of approaches applied to both AES and ASAS. Perhaps the oldest of these is known as the Bag of Words (BoW), which generally combines rules based on linguistic features in addition to a set of frequency-based features \cite{page_project_2003}. As Natural Language Processing (NLP) began incorporating neural network-based models, these models were applied to AES and ASAS. Early implementations of neural network-based scoring \cite{dong_attention-based_2017, ormerod_automated_2021} used layers of recurrent units such as the long-short-term memory (LSTM) unit \cite{hochreiter_long_1997} and gated recurrent units (GRU) \cite{cho_learning_2014} with attention \cite{graves_hybrid_2013}. 

The most influential change to NLP has been the rise of attention \cite{graves_towards_2014} and the transformer architecture \cite{vaswani_attention_2017}. The use of transformer-based Large Language Models (LLMs), such as BERT \cite{devlin_bert_2019}, to perform ATS is now well-established in both AES \cite{rodriguez_language_2019, uto_automated_2020, yang_enhancing_2020} and ASAS \cite{ormerod_short-answer_2022}. In the past few years, generative language models (GLM)s like ChatGPT \cite{openai_gpt-4_2023} have garnered immense excitement from both the media and academic circles. These GLMs are pretrained on a large corpus and then instruction-tuned to perform a multitude of tasks \cite{chung_scaling_2022}. 

Attempts at ATS with GLMs have focused primarily on large, proprietary models, e.g. \cite{mizumoto_exploring_2023}, which raises several concerns in an educational setting. Firstly, given that student data can include personally identifiable information, the reliance on an externally managed API poses a security risk. Secondly, since the weights are not publicly available, there is no ability to apply tools from explainable AI (xAI) \cite{linardatos_explainable_2021}. From the viewpoint of sustainability, closed-source models can require much more resources to run and can be much more expensive in the long run. Some researchers have explored AES and feedback using small, open-source models: In \cite{stahl_exploring_2024}, there is an exploration of prompting strategies and machine evaluation of feedback correlates with human evaluation of feedback; it is also clear, however, that with respect to AES, in-context GLM performance remains far below that of fine-tuned classification models. 

\subsection{Model-generated feedback}

If we limit our research into GLMs merely to improve existing scoring systems, then we will have missed out on the potential to enhance educational assessment. There is a growing call from educators, students, and other stakeholders for these models to be used to provide feedback. 

% Currently, there are several strands of research into model-generated feedback in the context of educational assessment. 

Although model-generated feedback holds potential value for educators, there remain substantial hurdles to producing feedback that is useful. These limitations revolve around the the quality of feedback itself, as well as the difficult endeavor of validating that the feedback is indeed useful in a given context. With respect to feedback quality, even large GLMs produce hallucinations. In the field of text generation, \emph{hallucination} refers broadly to text that, while grammatically correct, is also nonsensical, unfaithful, unreliable, inaccurate, irrelevant, etc. \cite{ji_survey_2023, ye_cognitive_2023}. With respect to validation, there is no methodology in the field that can be used to easily validate such feedback. There are, moreover, no easy-to-implement systems to capture feedback in an on-going way from educators, which makes development of process-oriented tools extremely challenging. 

Beyond technological limitations, there are social implications that need to be considered in the face of novel educational technologies. The Substitution Augmentation Modification Redefinition (SAMR) model for technological innovation and adoption in educational settings, for instance, has been critiqued for justifying hierarchical approaches to product development and implementation \cite{hamilton_substitution_2016}. Technological advances which are described or marketed as educational tools need to be developed in tandem with teachers, administrators, and other educational practitioners. Although much of the enthusiasm (as well as economic pressure) behind feedback generation is warranted, this cannot supersede the need for taking a rigorous and ethical approach towards researching and developing such tools.

\subsection{Architecture of Generative Language Models for the GPU Poor}

In contrast to the large, proprietary GLMs that have dominated public attention, there is a concomitant open-source movement that strives to makes GLMs accessible to all. These relatively small, open-source models are typically released in ~7Gb and ~70Gb versions by researchers who are often affiliated with the same organizations that develop proprietary GLMs. For instance, Google recently released Gemma, Meta released Llama-3, and Microsoft released Phi-3. In contrast to their large, proprietary counterparts, these GLMs can run on (and can even be trained on) consumer-grade hardware, such as a single 24Gb GPU. That is, these models can be leveraged by the ``GPU poor'', which includes most of us educational researchers. This open-source movement allows researchers to experiment directly with GLMs, and to explore targeted use cases in education. Researchers have just begun to explore smaller, open-source GLMs for ATS and feedback (e.g. \cite{stahl_exploring_2024}). 

Although performance generally increases with scale, smaller GLMs perform surprisingly well. GPT-4 and Claude are enormous, and it is no surprise that they dominate leaderboards, yet their smaller, open-source counterparts (which require only a fraction of the memory) are not far behind. One reason that smaller GLMs are not further behind is that, aside from small variations, they generally share the same architecture. Furthermore, within the current paradigm, there is a consensus among researchers that the primary bottleneck to increasing performance is data volume and quality, not model architecture.

Current SOTA GLMs use a decoder-only architecture, sometimes combined with Mixture of Experts. The underlying design is actually simpler than the original transformer architecture advanced in Bidirectional Encoder Representations from Transformers (BERT, \cite{devlin_bert_2019}). Following the advent of BERT, many researchers proposed variants of BERT that improved either the data \cite{peng_mathbert_2021}, architecture \cite{jiang_convbert_2020, sun_mobilebert_2020}, or training schemes \cite{clark_electra_2020, he_debertav3_2021} of the original model. These models were predominantly encoder-only models which were made into classifiers by replacing the linear layer that predicts masked tokens with another randomly initiated linear layer (i.e. the classification head). 
% The performance of these models was tested by fine-tuning the model and classification head on a collection of tasks known as GLUE \cite{wang_glue_2019}. 
Encoder transformer-based pretrained language models are typically given a classification head, where the loss function is cross-entropy (e.g., see \cite{rodriguez_language_2019}).\footnote{It is also possible, though less common, to use the single target variant with a mean-squared error loss function \cite{yang_enhancing_2020}.} Many previous authors have applied transformer-based language models to AES and ASAS in this way \cite{uto_neural_2020, rodriguez_language_2019, yang_enhancing_2020, ormerod_short-answer_2022, sung_pre-training_2019}. Indeed, this is the current paradigm in most of AES and ASAS.

While this paradigm (of affixing a classification head) could also be applied to GLMs,\footnote{Indeed, this w done with the first GPT model \cite{radford_improving_2018}} this disregards the relationship learned by the model between the linear layer that predicts tokens and the transformer layers. The final output layer, however, can be left as is, and fine-tuning can focus on the intermediate layers (e.g., using QLoRA \cite{dettmers_qlora_2023}, described below). Because this form of fine-tuning preserves the relationship learned by the model between the linear layer, the models themselves retain much of their abilities as generative models when applied to more general tasks. This allows the models to be further prompted to produce feedback where the scores are at least able to be validated against known human-defined targets. 

% Each model was first pretrained on a large corpus and then instruction-tuned, hence, we expect that the models respond appropriately to sequences in the format of an instruction. 
% Although GLMs have important individuals differences, they all share an underlying transformer-based model architecture. 
% We acknowledge that there are a large number of other models we could have tried in this range.

The rapid growth of large language models, now reaching hundreds of billions of parameters, has introduced considerable engineering challenges for their large-scale deployment. A primary concern is training these enormous models within memory constraints. Generally, each parameter and its gradient are stored in 32-bit precision, requiring 4 bytes per trainable parameter. Advanced optimizers such as Adam with weight decay further increase memory consumption by storing additional data for each parameter. For example, fine-tuning a model with 7 billion parameters would typically need at least 28GB of video memory, excluding context length. 

To get around the typical memory requirements of GLMs, we employ a combination of two approaches: (1) quantization \cite{dettmers_qlora_2023}, wherein parameters are stored at lower precision, and (2) Low-Rank Adapters (LoRA) \cite{hu_lora_2021}. The combination of these methods is commonly referred to as QLoRA \cite{dettmers_qlora_2023}. 
% Furthermore, this paper takes preliminary steps toward demonstrating the feasibility of a pipeline that extends beyond scoring to provide feedback. Most importantly, this pipeline can be entirely trained and run securely on consumer-grade hardware. 
% We use the fine-tuned models to also provide scoring rationale which can be used as feedback and analyze the feedback using a grounded approach. 
Quantization converts the model's parameters from 32-bit floats to 4-bit NormalFloat data types \cite{dettmers_8-bit_2022}. Memory savings are further increased through double quantization, where the quantization constants themselves are also quantized. Despite using less memory, quantized models generally maintain robust performance. Additionally, memory can be further conserved by using 8-bit optimizers, which store variance and its square in 8-bit precision \cite{dettmers_8-bit_2022}.
Low-rank adaptation (LoRA, \cite{hu_lora_2021}), is an increasingly popular method of parameter-efficient fine-tuning \cite{xu_parameter-efficient_2023}. In the following section, we describe LoRA in detail.

\subsection{Training Generative Language Models for the GPU Poor}

LoRA is a powerful, parameter-efficient technique for fine-tuning GLMs. In combination with quantization, it makes it possible to fine-tune GLMs using less than 8Gb of memory, thereby making them more feasible for development and deployment.

The central idea behind LoRA is that we seek to update the large feed-forward layers of the model by only considering a low-rank additive component, initially set to 0. Mathematically, we suppose a linear layer is represented by
\[
L(x) = W_0 x + b
\]
where $W_0 \in \mathbb{R}^{d\times k}$ is the original pretrained weight matrix and $x$ is the input. It is known that updates to the linear transformations are sparse and in many cases, approximated well by matrices of low-rank. We seek to update the weight matrix, $W \to \tilde{W}$ in a single step by
\[
\tilde{W} = W_0 + \delta W = W_0 + BA 
\]
where $A \in \mathbb{R}^{r\times k}$ and $B \in \mathbb{R}^{d\times k}$. 

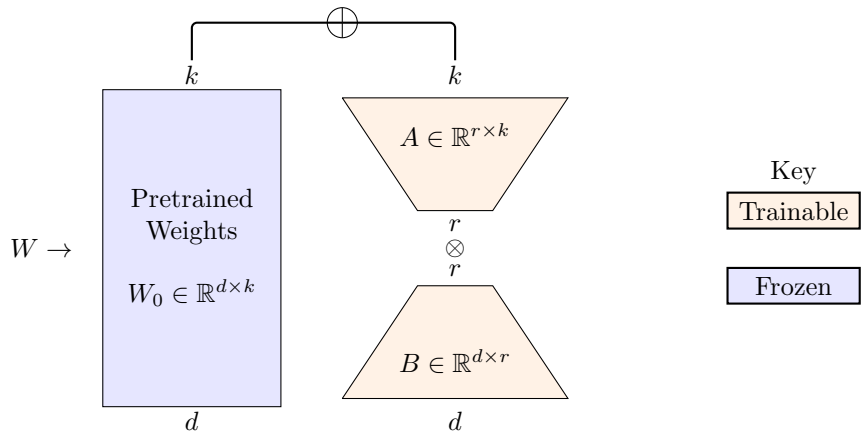
\begin{figure}
\begin{center}
\begin{tikzpicture}[]
\node[fill=blue!10 ,minimum height = 12em, draw=black] at (0,0) {\begin{tabular}{c}Pretrained\\ Weights\\ \\ $W_0 \in \mathbb{R}^{d\times k}$ \end{tabular}};

\draw[fill=orange!10 ,minimum height = 10em] (2,-2) -- (5,-2) -- (4,-.5) -- (3,-.5) -- (2,-2);
\draw[fill=orange!10 ,minimum height = 10em] (2,2) -- (5,2) -- (4,.5) -- (3,.5) -- (2,2);
\node at (3.5,-1.5) {$B\in \mathbb{R}^{d\times r}$};
\node at (3.5,1.5) {$A\in \mathbb{R}^{r\times k}$};
\node at (3.5,0) {$\otimes$};
\node at (3.5,.3) {$r$};
\node at (3.5,-.3) {$r$};
\node at (3.5,-2.3) {$d$};
\node at (0,-2.3) {$d$};
\node at (0,2.3) {$k$};
\node at (3.5,2.3) {$k$};
\draw (2,3) circle (2mm);
\draw (2,3-.2) -- (2,3+.2);
\draw[rounded corners=2pt, thick] (0,2.5) |- (2,3);
\draw[rounded corners=2pt, thick] (3.5,2.5) |- (2,3);
\node[fill=orange!10, draw=black, thick, minimum width=5em] at (8,0.5) {Trainable};
\node[fill=blue!10, draw=black, thick, minimum width=5em] at (8,-0.5) {Frozen};
\node at (8,1) {Key};
\node at (-2,0) {$W \to $};
\end{tikzpicture}
\end{center}
\caption{A visual representation of the training scheme for LoRA. The update of $W$ is given by $W_0 + BA$ where $B$ and $A$ are of a particular low-rank form.}
\end{figure}

In this setting, it is expected that $r << \min(d,k)$ so that the number of trainable parameters is $r(k+d)$. Typical values of $r$ (e.g., $2< r < 32$) are chosen such that the number of trainable parameters is far fewer than full-parameter fine-tuning. 

The advantages of LoRA include reduced memory requirements for saving fine-tuned models, more efficient training, no impact on inference speed, and the capacity for combination with other parameter efficient fine-tuning methods. The memory requirements for saving a fine-tuned large language model with LoRA are limited to size of the pairs of update matrices, which orders of magnitude smaller than the original model. Training is also more efficient and requires less GPU memory since gradients only need to be calculated for the update matrices. The impact on inference latency can be reduced to zero if the update matrices are added to the pretrained weights and subsequently removed from the model after loading. Finally, because the update matrices can be removed, LoRA can be combined with any other adapters \cite{poth_adapters_2023}.

\section{Methods}\label{sec:method}

% In this section, we review characteristics of the data for both the AES and ASAS datasets. We detail what models we use, and explicate training procedures, including techniques and prompts used to fine-tune the models in a parameter-efficient manner.

\subsection{Data}

The Automated Student Assessment Prize (ASAP) AES and SAS datasets were originally made available to the public via two competitions hosted by Kaggle in 2012 \cite{shermis_state---art_2014}. The AES dataset encompasses a total of 12,978 essays, spanning 8 distinct stimuli.\footnote{We use the term \emph{stimuli} or \emph{items} instead of \emph{prompts}, as the latter is easily confused with \emph{prompts} used to query GLMs.} The SAS dataset consists of 17,043 total responses across 10 items that span various subjects, administered to students in grades 8 and 10 (depending on the item). Each response was scored by two human annotators. Accompanying the scored data are comprehensive scoring rubrics that include scoring guidelines and score ranges tailored to each stimulus. One of the advantages of using the AES and SAS datasets are that they are commonly used by other researchers, allowing us to compare our results with a wide range of previously established approaches. 

In order to maintain comparability with the extensive literature on these datasets, test-train splits were chosen to align with previous studies \cite{taghipour_neural_2016, dong_attention-based_2017, rodriguez_language_2019, uto_automated_2020, yang_enhancing_2020, ormerod_argumentation_2023, ormerod_short-answer_2022, kumar_automated_2021, kumar_get_2019}. For the AES dataset, we follow the five-fold cross-validation defined by \cite{taghipour_neural_2016}. For the SAS dataset, we used the same splits used in previous studies (e.g., \cite{ormerod_short-answer_2022, kumar_get_2019}. The (average) size of the training, development (or dev), and test sets for the AES and SAS datasets, in addition to some basic characteristics of the datasets, are presented in Table \ref{tab:IRRstats}

\begin{table}[!ht]
    \centering
    \begin{small} 
    \begin{tabular}{l r ccccccc} \toprule
    & & train & dev & test & \multicolumn{2}{c}{Test} & Avg. Wrds. / & Score\\
    &   Set & N & N & N & QWK & Acc & Response & Range \\ \midrule
Essay   &1 & 1070& 357 & 357 & .721 & .653 & 366 & 2-12\\
    &    2 & 1080& 360 & 360 & .814 & .783 & 381 & 1-6\\
    &    3 & 1036& 345 & 345 & .769 & .749 & 109 & 0-3\\
    &    4 & 1063& 354 & 354 & .851 & .772 & 94 & 0-3\\
    &    5 & 1083& 361 & 361 & .753 & .580 & 122 & 0-4\\
    &    6 & 1080& 360 & 360 & .776 & .623 & 154 & 0-4\\
    &    7 & 941& 314 & 314 & .721 & .292 & 168 & 2-24\\
    &    8 & 434& 145 & 145 & .624 & .278 & 605 & 10-60\\ 
    \midrule
Short    &    1 & 1002 & 335 & 335 & .938 &  .904 & 52 & 0-3\\
Answer    &    2 & 1278 & 256 & 256 & .911 & .848 & 65 & 0-3\\
    &    3 & 1084 & 362 & 362 & .762 & .785 & 53 & 0-2\\
    &    4 & 993 & 332 & 332 & .686 & .783 & 46 & 0-2\\
    &    5 & 1077 & 359 & 359 & .935 & .958 & 28 & 0-3\\
    &    6 & 1077 & 360 & 360 & .973 & .967 & 28 & 0-3\\
    &    7 & 1079 & 360 & 360 & .968 & .958 & 46 & 0-2\\
    &    8 & 1079 & 360 & 360 & .837 & .833 & 60 & 0-2\\
    &    9 & 1078 & 360 & 360 & .831 & .808 & 54 & 0-2\\
    &    10 & 984 & 328 & 328 & .904 & .909 & 45 & 0-2\\ \bottomrule
    \end{tabular}
    \end{small}
    \caption{Characteristics of ASAP AES and SAS datasets.}
    \label{tab:IRRstats}
\end{table}

The scoring rubric for the AES dataset emphasizes proper spelling and grammar usage, logical organization with smooth transitions between ideas, and the ability to exhibit analytical comprehension backed by supporting evidence. The rubrics for essay set 1,7, and 8 do this by breaking the score into several traits. The final score is the sum of each of the trait scores. While some of the essay topics depend on a particular prompt, the rubric can be generally interpreted independently of any prompt.

In contrast, the rubrics for the SAS items focus on specific pieces of information that need to be in a response in order to obtain a score. These short answer questions are designed to test knowledge and comprehension, hence grammar and spelling are not a part of the rubric.  

\subsection{Performance Metric}

When evaluating the model performance, we compute Quadratic Weighted Kappa (QWK), which was the original metric specified in the Kaggle competitions \cite{shermis_state---art_2014, shermis_contrasting_2015}. A rough interpretation of this metric is that it measures the probability above chance that two raters agree: a QWK of $1$ indicates exact agreement, $0$ indicates random agreement, and $-1$ indicates perfect disagreement. This metric is also standard in the industry for comparing machine scoring performance \cite{williamson_framework_2012}. 

\subsection{Models}

In selecting models for our study, we prioritized those that could operate on standard consumer hardware while still delivering performance adequate for generating useful feedback. We identified four models that met these criteria and represented the forefront of open-source model development from major contributors in the field. These include (with affiliation in parentheses): Llama-3 (Meta), Mistral v0.2 (Mistral), Gemma-1.1 (Google), and Phi-3 (Microsoft). Table \ref{tab:model_characteristics} provides a brief overview of architectural characteristics, along with the total parameter count and references to their respective technical documentation. 
% The fundamental architecture of these models is characterized by several key factors: the number of layers, hidden size, intermediate size, and vocabulary size. 

\begin{table}[!ht]
    \centering
    \begin{tabular}{lcccccc} \toprule
     Model & Release & \# & \# & Hidden & Intermediate & \# Vocab.\\ 
     Name & Date & Params. & Layers & Size & Size & Tokens\\ 
     \midrule
     Mistral v0.2-Instuct \cite{jiang_mistral_2023} & 12/11/2023 & 7.24B & 32 & 4,096 & 14,336 & 32k\\
     Gemma 1.1-Instruct \cite{gemma_team_gemma_2024} & 3/26/2024 & 8.54B  & 28 & 3,072 & 24,576 & 256k\\
     Llama-3-8B-Instruct \cite{meta2024introducing} & 4/17/2024 & 8.03B & 32 & 4,096 & 14,336 & 128k\\
     Phi-3-7B-Instruct \cite{abdin_phi-3_2024} & 4/22/2024 & 3.82B & 32 & 3,072 & 8,192 & 32k\\ \bottomrule
    \end{tabular}
    \caption{Model characteristics. Note: Original Release date determined by date of original commit on huggingface-hub.\label{tab:model_characteristics}}
\end{table}

One model was trained for each item, resulting in a total of 40 trained models (4 model types x 10 items). 

\subsection{Parameter-efficient fine-tuning}

Models were loaded through Huggingface-hub, quantized into smaller, 4-bit models using bitsandbytes, and trained using low-rank adaptors (LoRA). Learning rate was set to 2e-4 (except for Gemma-1.1, which was set to 1e-4 to ensure convergence), with a linear rate decay over 10 epochs. $r$ and $\alpha$, key parameters for LoRA, were each set to 32. Table \ref{tab:lora-characteristics} lists how this $r$ value affects trainable parameters and memory used for each of the four models. 

\begin{table}[!ht]
    \centering
    \begin{tabular}{l|ccc|cc} \toprule
         & & \# Trainable & Memory & Training & Inference \\
        Model & r-value & Params. & Used & Time & Time \\ 
        \midrule
     Mistral v0.2-Instuct & 32 & 83.9M &  4.67Gb & 10.8 & 30.7 \\
     Gemma 1.1-Instruct & 32 & 100M & 6.01 Gb & 12.4 & 37.6 \\ 
     Llama-3-8B-Instruct & 32 & 83.9M & 5.76Gb & 10.5 & 29.8 \\
     Phi-3-7B-Instruct & 32 & 59.8M &2.40Gb & 6.6 & 18.4 \\ 
     \bottomrule
    \end{tabular}
    \caption{Size of models in terms of trainable parameters, memory requirements, and training and inference times.}
    \label{tab:lora-characteristics}
\end{table}

To ease GPU load, training data were not batched (i.e. batch size was 1), and context length was capped at 2,048 (note that this cap was not exceeded for any response). We used an early stopping criterion, based on best QWK performance on the development set, computed at the end of each epoch, within a span of 10 epochs. Models were trained on a 24GB A10 GPU. 

We calculated training and inference times of each model. Times were transformed so as to be relative to the training and inference times of a standard BERT-base classification model. Thus, for example, Mistral took 10.8 times longer to train than BERT, and 30.7 times longer to predict scores on the test set. The BERT model was trained in batches of 4, over the span of 20 epochs, and on the same hardware as the GLMs. 

\subsection{Prompting for Score Prediction}

We used the following template to prompt the model for a score, given an item-specific max score, an item-specific rubric, and a student response (all indicated by curly brackets below). Note that “User” and “Assistant” role formats vary between models; roles were not entered into the prompt itself, but handled automatically via Huggingface’s apply\_chat\_template function. 

\begin{quote}

\textbf{User} You are a grading assistant. Assign a **Score** between 0 and \{max\_score\} using the **Rubric** provided to a **Student Response**
\\\\
**Rubric**
\\
\{item\_rubric\}
\\\\
**Student Response**
\\
\{student\_response\}
\\\\
\textbf{Assistant} Score: 

\end{quote}

Using the filled-out template as input, we constrained the model to generate one additional token. If the model generated a non-integer token, then the score was given a 0. 

\subsection{Prompting for Feedback Generation}

After prompting for score predictions, we incorporated the predicted scores into another template to prompt the model for feedback generation. Although much of the feedback generation template is identical to the score prediction template, the model was prompted separately. A maximum of 256 new tokens were produced for AES feedback and 128 tokens for SAS.

\begin{quote}

\textbf{User} You are a grading assistant. Assign a **Score** between {min\_score} and \{max\_score\} using the **Rubric** provided to a **Student Response**
\\\\
**Rubric**
\\
\{item\_rubric\}
\\\\
**Student Response**
\\
\{student\_response\}
\\\\
\textbf{Assistant} Score: \{predicted\_score\}
\\\\
\textbf{User} Using the rubric, specify why you gave the response a score of \{predicted\_score\}.
\\\\
\textbf{Assistant}\footnote{This last Assistant Prompt was only included for short answer items} The response was given a score of \{predicted\_score\} because 

\end{quote}

\subsection{Qualitative Analysis of Feedback}

To characterize the differences in feedback provided by each of the 4 models, we sampled student responses with predicted scores that matched human rater scores. For the SAS dataset, we sampled responses across all possible score points for 2 science items (Items 1 and 10) and 2 ELA items (Items 3 and 7). We analyzed 13 student responses across 4 items (and 2-3 possible score points), for a total of 52 explanations. For the AES dataset, we sampled responses across all possible score points for 2 stimuli (Items 2 and 3). We analyzed 10 student responses across 4 items (and 4-6 possible score points) for a total of 40 explanations. 

In analyzing responses, we took a grounded approach (Creswell and Poth, 2016 -- add citation). The philosophy behind grounded qualitative research is to let patterns emerge from the data, rather than approach the data with pre-defined codes or hypotheses. More specifically, analyses consisted of two phases. In the first phase, we read through responses, noted salient trends, summarized notes, and revisited notes for each response. In the second phase, we summarized these notes into general patterns and trends, and identified consistent and inconsistent examples in the data.

\section{Results}\label{sec:results}

Results are divided into four section: In sections 1 and 2, we present the results of fine-tuned GLMs on AES and ASAS, respectively; in sections 3 and 4, we characterize feedback after prompting GLMs to explain their scores based on item-specific rubrics, for AES and ASAS, respectively. 
% Since these two datasets have been used extensively, we provide only the most relevant benchmarks for comparison, including the use of a BoW/rules-base approach, traditional recurrent network, some language model, and the state-of-the-art. We also provide baselines given by the models with no fine-tuning. 

\subsection{Automated Essay Scoring}

Table \ref{tab:aes_results} presents the results of fine-tuned GLMs on performing AES on the ASAP-AES datset. We provide comparisons to several notable benchmarks pertinent to the task. These include the original human-human agreement score \cite{shermis_state---art_2014}, the BoW results reported in \cite{taghipour_neural_2016} and subsequent modifications using attention mechanisms \cite{dong_attention-based_2017}, the original BERT results \cite{rodriguez_language_2019}, the current SOTA performance \cite{xie_automated_2022}, ``fine-tuned" GPT-3.5 \cite{mansour_can_2024}, and GPT-4 \cite{xiao_human-ai_2024}. In addition to these important reference points, we also provide results from off-the-shelf, i.e. not fine-tuned, models (no asterisks) alongside fine-tuned models (indicated with asterisks). 

\begin{table}[!ht]
\begin{small}
\begin{center}
\begin{tabular}{l |  cccccccc | c} \toprule
Model & 1 & 2 & 3 & 4 & 5 & 6 & 7 & 8 & Avg.\\ \midrule
Human \cite{shermis_state---art_2014} & .721 & .812 & .769 & .850 & .753 & .775 & .720 & .620 & .752 \\ \hline
EASE \cite{taghipour_neural_2016} & .781 & .621 & .630 & .749 & .782 & .771 & .727 & .534 & .699 \\
LSTM+CNN+Att \cite{dong_attention-based_2017} & .822 & .682 & .672 & .814 & .803 & .811 & .801 & .705 & .764 \\ 
BERT (base) \cite{rodriguez_language_2019} & .792 & .680 & .715 & .801 & .806 & .805 & .785 & .596 & .758 \\
NPCR \cite{xie_automated_2022} & .856 & .750 & .756 & .851 & .847 & .858 & .838 & .779 & .817\\
GPT-3.5* \cite{mansour_can_2024} & .741 & .618 & .704 & .859 & .796 & .848 & .727 & .614 & .738 \\ 
GPT-4 \cite{xiao_human-ai_2024} & .280 & .338 & .331 & .784 & .623 & .728 & .257 & .454 & .474 \\
\hline
Mistral-7B-Instruct-v0.2         & .595 & .359 & .583 & .740 & .497 & .460 & .320 & .060 & .452\\
Mistral-7B-Instruct-v0.2*        & .831 & .702 & .695 & .833 & .822 & .818 & .830 & .728 & .782\\
Gemma-1.1-7b-it                  & .214 & .516 & .427 & .361 & .251 & .376 & .425 & .293 & .358\\
Gemma-1.1-7b-it*                 & .809 & .711 & .688 & .826 & .802 & .818 & .824 & .623 & .763\\
Meta-Llama-3-8B-Instruct         & .255 & .463 & .432 & .557 & .653 & .608 & .283 & .362 & .452\\
Meta-Llama-3-8B-Instruct*        & .821 & .727 & .717 & .824 & .815 & .829 & .837 & .752 & .789\\
Phi-3-mini-4k-instruct           & .408 & .334 & .299 & .465 & .605 & .557 & .279 & .269 & .402\\
Phi-3-mini-4k-instruct*          & .827 & .714 & .715 & .828 & .830 & .827 & .837 & .710 & .786 \\ \bottomrule
\end{tabular}
\end{center}
\end{small}
\caption{The results of fine-tuning on the ASAP AES dataset. The models that were fine-tuned are labeled with an *. \label{tab:aes_results}}
\end{table}

The fine-tuned generative models performed well compared to standard benchmarks. They exceeded performance of AES, BERT (base), fine-tuned GPT-3.5, and the combination of LSTM, CNN, and attention mechanisms. Although none of the models achieve the current SOTA performance (a distinction held by NPCR), each individual model surpasses many previous benchmarks. Fine-tuned GLMs also seem comparable, if not above, human-level performance.\footnote{Regarding comparability to human-human QWK, it should be noted that the models were trained on the \emph{resolved scores}, which have different ranges than the original human scores. According to the rubric, the resolved scores are calculated as the sum of the two human scores for items 1, 7, and 8.}

\subsection{Automated Short Answer Scoring}

The performance of GLMs fine-tuned for ASAS are presented in Table \ref{tab:asas_results}. Fine-tuned models are indicated with asterisks. As with AES, there are a number of important results in the literature to compare against our own results. Firstly, there is the human agreement score \cite{shermis_contrasting_2015}, the rule-based approach known as AutoSAS \cite{kumar_get_2019}, the current SOTA given by an ensemble of pretrained models \cite{ormerod_short-answer_2022}, ``fine-tuned" GPT-3.5 \cite{chamieh_llms_2024}, and GPT-4 \cite{jiang_short_2024}. Results from non-fine-tuned versions of each of the 4 models (no have asterisks) are also included. 

\begin{table}[!ht]
\begin{small}
\begin{center}
\begin{tabular}{l |  cccccccccc | c} \toprule
Model & 1 & 2 & 3 & 4 & 5 & 6 & 7 & 8 & 9 & 10 & Avg\\ \midrule
Human \cite{shermis_contrasting_2015} & .938 & .911 & .758 & .686 & .935 & .973 & .968 & .837 & .831 & .904 & .874 \\ \hline
AutoSAS \cite{kumar_get_2019} & .872 & .824 & .745 & .743 & .845 & .858 & .725 & .624 & .843 & .832 & .791\\
BERT-base & .849 & .772 & .692 & .722 & .845 & .840 & .676 & .598 & .829 & .717 & .749 \\ 
Ensemble LLM \cite{ormerod_short-answer_2022} & .882 & .891 & .722 & .750 & .813 & .822 & .734 & .702 & .865 & .779 & .796 \\
GPT-3.5* \cite{chamieh_llms_2024} & & & & & & & & & & & .610 \\
GPT-4 \cite{jiang_short_2024}  & .715 & .724 & .626 & .517 & .772 & .799 & .495 & .553 & .703 & .865 & .677 \\
\hline
Mistral-7B-Instruct-v0.2 & 0.57 & .449 & .188 & 0.33 & .331 & .496 & .243 & .280 & .507 & .529 & .392 \\
Mistral-7B-Instruct-v0.2* & .864 & .807 & .725 & .636 & .776 & .855 & .746 & .697 & .771 & .709 & .759 \\
Gemma-1.1-7b-it & .315 & .341 & .042 & .087 & .194 & .307 & .168 & .195 & .164 & .522 & .233 \\
Gemma-1.1-7b-it* & .859 & .814 & .602 & .659 & .824 & .798 & .757 & .716 & .751 & .712 & .749 \\
Meta-Llama-3-8B-Instruct & .427 & 0.45 & .293 & .334 & .602 & .563 & .188 & .361 & .420 & .615 & .425 \\
Meta-Llama-3-8B-Instruct* & .878 & .823 & .687 & .649 & 0.82 & .807 & .714 & .659 & .759 & .733 & .753 \\
Phi-3-mini-4k-instruct & .452 & .360 & .157 & .281 & .341 & .449 & 0.36 & .126 & .395 & .397 & .332 \\
Phi-3-mini-4k-instruct* & .864 & .779 & .697 & .625 & .781 & 0.85 & .718 & .691 & .788 & .703 & .750 \\
\bottomrule
\end{tabular}
\end{center}
\end{small}
\caption{The results of fine-tuning on the ASAP-SAS dataset. The models that were fine-tuned are labeled with an *.\label{tab:asas_results}}
\end{table}

In contrast with AES, the results of pertaining these large models offers comparable, but not superior, performance to BERT. The GLMs seem do outperform previous benchmarks on items 7 and 8; the results for Gemma and Mistral are above previously known models \cite{ormerod_short-answer_2022}. The performance on items 4 and 9, however, are lower than the benchmarks provided. 

\subsection{Automated Feedback for Essay Scoring}

After GLMs predicted scores, we prompted them for feedback---in this case, an explanation for the score based on the scoring rubric. To illustrate the type of feedback generated by each of the four models, we present the feedback generated in response to an essay on item 1 (Table \ref{tab:aes_explanation}). The essays was assigned a score of 8 by all GLMs.

\begin{table}
\begin{tabular}{p{.15\textwidth} | p{.85\textwidth}} \toprule
{\bf Model} & {\bf Feedback} \\ 
\midrule
Mistral-7B-Instruct-v0.2 & I gave the response a score of 8 because it was well-written, organized, and persuasive. The author made a strong argument about the benefits of computers and provided specific details and examples to support their position. They also acknowledged potential counter-arguments and addressed them in a thoughtful and respectful manner. Overall, the response was convincing and persuasive. \\ \hline
gemma-1.1-7b-it & Your response is well-written, clear, and easy to read. You have a strong understanding of the topic and you have supported your reasons well. You have also used persuasive language to convince the reader to agree with you. I would like to add that I think you could have added a little more detail to your reasons, but other than that I think you have done a good job.\\ \hline
Meta-Llama-3-8B-Instruct & 
Here's why I gave the response a score of 8:

Strengths:
\begin{itemize}
\item The response is well-organized and easy to follow.
\item The writer provides specific details and examples to support their ideas.
\item The writer shows a good understanding of the topic and the audience.
\end{itemize}
Weaknesses:
\begin{itemize}
\item The response is not as polished as it could be. There are some awkward phrasings and minor errors.
\item The writer could have provided more elaborate explanations and more specific details to support their ideas.
\item The response lacks a strong, attention-grabbing opening that draws the reader in.
\end{itemize}

Score: 8

I gave the response a score of 8 because it is well-organized and easy to follow, but it is not as polished as it could be. The writer provides some specific details and examples to support their ideas, but could have provided more elaborate explanations and more specific details. Overall, the response shows a good understanding of the topic and the audience, but could be improved with more attention to detail and more elaborate explanations.\\ \hline
Phi-3-mini-4k-instruct & I gave the response a score of 8 because it clearly takes a strong position on the effects of computers, providing well-elaborated reasons and specific details. The response exhibits strong organization, uses sophisticated language, and shows awareness of the audience. However, it could have provided more elaboration on the benefits of using computers for learning and communication, and could have included more specific details to further support the argument. Overall, it is a strong response but with a few areas for improvement.\\ \bottomrule
\end{tabular}
\caption{Feedback from each model for the first response in the dataset. \label{tab:aes_explanation}}
\end{table}

By examining the feedback across items, responses, and models, we found that the feedback provided by fine-tuned versions of Mistral and Gemma tended to be more repetitive as the models seemed to settle into a loop more readily than Phi-3 and Llama-3. For stimuli where the rubric relied on external information, such as the understanding of a text, the language models struggled to produce sensible feedback and often only summarized and reiterated aspects of the response, rather than detailing why the score was assigned.\footnote{It is worth noting that the stimuli were very long and including the stimuli in addition to the full rubric would have exceeded the context limits we imposed for practical considerations. Secondly, in the case that the resolved score was the addition of the trait scores for each rater, the rubric described only the rater score, not the resolved score. So we employed a language model to summarize the differences between a high and low-scoring essay. Perhaps managing this better could lead to more constructive feedback.} The models seem to provide much clearer feedback when the rubric could be interpreted independently of the stimuli (i.e. 1, 2, 7, and 8).

The most useful feedback overall seemed to come from fine-tuned versions of the Phi-3 and Llama-3 models. Even though they provided the most accurate explanations, they were not immune from repetition or errors.  

\subsection{Automated Feedback for Short Answer Scoring}

In Table \ref{tab:asas_explanation}, we present feedback from for a 1-point response to Item \#10. We selected this particular response because model feedback was typical of what we observed for other items and score points. 

For Item \#10, to get full credit (2 points), the student had to (1) “describe how [a chosen color] might affect the inside of the doghouse” and (2) “use results from the experiment to support [their] description.” The student response for this particular example reads, “black.  it might effect it,by using this color it can make the doghouse more warmer on summer days” (Id: 26865). The response does state that the color black would make the doghouse warmer (1 point), but fails to reference the experiment (0 points). Because it met 1 of the 2 criteria outlined in the rubric, it received a score of 1. Table \ref{tab:asas_explanation} provides the explanations given by each of the 4 models.

\begin{table}[!ht]
\begin{small}
\begin{center}
\begin{tabular}{p{.15\textwidth} | p{.8\textwidth}} 
\toprule
\textbf{Model} & \textbf{Explanation} \\ 
\midrule
Mistral-7B-Instruct-v0.2 & 1 is the lowest number on the chart about the lamp and color of the lid it had an average temperature of 42(DEG)C. \\ 
\hline
gemma-1.1-7b-it & 1: The doghouse would not be warm enough for the dog if it was painted white. 2: The experiment was not designed to see if white would be too cold for the dog. The average temperature for the white lid was 42(DEG)C, and for the black lid it was 53(DEG)C. The white lid was 11(DEG)C cooler than the black lid, and if the doghouse was painted white, it would not be warm enough for the dog. \\
\hline
Meta-Llama-3-8B-Instruct & 1) the color black was used and 2) the phrase ``it might effect it" was used but the response was incomplete and didn't provide enough information to determine the temperature of the inside of the doghouse. The response was also unclear about what the color black would do to the temperature of the inside of the doghouse. For example, would it make it warmer or colder? The response didn't provide enough information to determine the temperature of the inside of the doghouse. Therefore, the response was given a score of 1. \\
\hline 
Phi-3-mini-4k-instruct & 1 point were awarded for the response because the response chose a color and described how that color might affect the inside of the doghouse BUT the results from the experiment were missing. \\ 
\bottomrule
\end{tabular}
\end{center}
\end{small}
\caption{Example of explanations for short-answer scoring. Note: All model responses began with, “The response was given a score of 1 because ”, to help them produce sensical feedback.\label{tab:asas_explanation}}
\end{table}

Mistral did not produce an explanation for the score. Rather, it seemed to summarize part of the item stem, or perhaps it generated its own (student-like) response. It was common for Mistral to generate its own responses, which it would score, and subsequently produce another response and another score, and so on in a loop (not shown here).

In the above example, Gemma seems to have produced a (student-like) response, and provides no explicit reference to the rubric. The response is separated into two, however, which may indicate some kind of pastiche, blending a response with the form of the rubric. Although not evident in this example, Gemma tended to summarize or repeat student responses in its explanations. These summaries were sometimes accompanied by relevant aspects of the rubrics. 

In contrast to Mistral and Gemma, Llama-3 referenced the student response in an evaluative way. It mentioned the color chosen by the student, and it quoted a phrase from the response (“it might effect it”) that could impact its score. At the end, Llama-3 summarized its explanation with a definitive, “Therefore, the response was given a score of 1,” as if it had produced a satisfying justification. Yet there are two serious flaws in Llama-3’s explanation. First, it included statements that contradict the student response, i.e., the response \emph{was not} “unclear about what the color black would do to the temperature,” as Llama-3 claimed. And second, it omitted one of the criteria in the rubric (i.e. referencing the experiment), and entirely fabricated another in its place (i.e. the color does not have to be black, as implied). Although the explanation is appropriate in style, contains evaluative language, and references the student response, it misrepresents the rubric and the response. This was common of Llama-3 explanations, which were often odd combinations of the rubric and summaries of students’ responses. 

Lastly, Phi-3 provided a succinct and accurate explanation of why the student would receive a 1 for this response. Phi-3 was not infallible, but it often evaluated student responses with some justification of the score or explicit reference to the rubric. 

\section{Discussion}
\label{sec:discussion}

\subsection{Summary}

In this paper, we have demonstrated that it is possible to fine-tune small, open-source GLMs to (1) achieve adequate performance for AES and ASAS and (2) generate appropriate rationales (at least in some cases) for predicted scores. Our method pushes beyond the paradigm of appending a classification head to a pretrained language model, yet avoids the many issues involved in querying large, proprietary GLMs via APIs. We find that parameter-efficient fine-tuning (using no more than a 24Gb GPU) for relatively small, open-source GLMs exceeds performance of proprietary GLMs that are orders of magnitude larger. Furthermore, due to the efficient nature of training checkpoints, the only parameters that are required to serve these models are the LoRA weights, which amount to less than 100 million parameters, fewer parameters than a BERT model. Given the widespread enthusiasm and fear around GLMs, it may come as a surprise that they did not lead to SOTA results. Ensembles of smaller LMs remain more efficient and performant than GLMs for AES and ASAS. 
% \footnote{Traditionally, ASAS is a more difficult task, since rubrics are focused on information and content, as opposed to grammatical conventions, which are common in essay scoring rubrics. Traditional language modeling methods (e.g. Bag of Words), rely on $n$-grams that can be brittle when it comes to inconsistencies like spelling errors. It is interesting that this difficulty persists even in the era of GLMs.}

% Other researchers have found similar results, even with the likes of GPT-4 (BEA2023 paper from Duo). 

One of the unique advantages of using GLMs is the ability to move beyond scoring alone---in this study, we prompt the fine-tuned models to provide an explanation of the score. We found that models were capable of (sometimes) generating adequate justifications, and that Phi-3 was more consistent than the other models. Yet this study does not undertake a thorough analysis of model-generated feedback. Although preliminary results are encouraging, rigorous analysis is needed. This would include carefully defined constructs of interest, collaboration with educators and trained human raters, and targeted use cases that identify whom the feedback is for, when the feedback should be provided, and what shortcomings need to be avoided. It is noteworthy, however, that fine-tuned GLMs were able to generate feedback at all, especially given that they were fine-tuned to predict scores (i.e. not feedback). It has been shown that, even with some a small amount of fine-tuning, model behavior can change dramatically \cite{qi2023fine}. 

The performance of the GLMs explored in this study are promising, particularly since they avoid the critical issues of proprietary models. Firstly, these models can be run securely and efficiently with relatively low requirements. Although security is not a concern when examining performance on a publicly-available dataset, it is a concern in many educational contexts, where personally identifiable information about students may be shared with the organization hosted the GLM. Secondly, in order to interpret the output of these models, we must be able to access the weights. The lower computational requirements of smaller, open-source models allows them to be more readily used in explainable AI workflows. Thirdly, we believe that GLMs used for educative tasks should be developed by educators and educational researchers. The open-source movement in AI permits some agency in developing these tools, without relegating decisions to a few tech-focused companies. The methods prescribed in this paper can be duplicated without recourse to industrial-scale compute power. 
% It stands to reason that larger, open-source models would be able to produce better results (such as the larger variants of Llama-3, Mistral, and Phi-3), and we would also expect this of fine-tuned proprietary models, were they accessible or affordable to train. 

\subsection{Comparison to Proprietary GLMs}

With respect to scoring, our fine-tuned results far exceed those of ``fine-tuned" GPT-3.5 for both AES \cite{mansour_can_2024} and ASAS \cite{chamieh_llms_2024}. We put ``fine-tuned" in quotation marks because the fine-tuning procedure(s) available to the public are undisclosed and optimization (e.g. modulating the learning rate) is not currently available. Given that GPT-3.5 is vastly larger in size (175B) and requires far more computation \cite{brown2020language} compared to the models explored in our study, it is surprising that its performance is so underwhelming. Our results are also superior to (non-fine-tuned) GPT-4 with respect to both AES \cite{xiao_human-ai_2024} and ASAS \cite{jiang_short_2024}. It should be noted that fine-tuning is not currently available for GPT-4; yet even if fine-tuning were available and results were adequate, these would be subject to the same limitations outlined above. 
% Our results further underscore the limitations of relying on large, proprietary GLMs to perform educative tasks: Not only are there issues around privacy, security, transparency, efficiency,  In addition to the weaknesses described throughout, they are also underperformant. 
We note that our study does not undertake a comparison of feedback between large, proprietary GLMs and smaller, open-source GLMs; it may be that large GLMs excel in this area.

\subsection{Limitations}

As noted previously, this study does not attempt to provide quantitative empirical evidence regarding the validity of model-generated feedback. Model-generated feedback, although promising, requires more rigorous evaluation that should be undertaken in collaboration with educational practitioners. Even for the relatively humble task of providing an explanation for a score, models were far from infallible. More research is needed to validate that the model is consistently connecting scores to the rubric. There are others who are exploring the more complicated task of producing model-generated feedback that is useful to educational practitioners (e.g. \cite{stahl_exploring_2024, xiao_human-ai_2024}). Robust feedback systems likely require on-going evaluation, and may depend on human-in-the-loop frameworks. 

Although there is growing pressure to develop educational tools using GLMs, there is no easy method of validating feedback. At this stage, the validation of feedback should be a primary concern for the future for the use of GLMs in education. This may mean the creation of datasets that are focused on feedback, or the use of existing information, such as essay trait scores, to validate existing feedback. To help facilitate such analyses, We have open-sourced the feedback provided on a single validation sample in the hopes of prompting further analyses\footnote{https://github.com/christopherormerod/kaggle\_aes\_asas\_feedback}. One thing that is fairly clear at this stage is that these models are computationally capable of being used in such a pipeline. The question remains, however, as to whether they are valid for carefully defined, targeted use cases.

% Although the above sampling procedure allows for more fair comparisons, it does not compare feedback from responses that received different scores; these responses may be different (e.g. harder to score), and may reflect different types of feedback that were not considered. 

% \subsection{Future Research}

% There are several interesting avenues for future research. First, as mentioned above, there is great deal of research that needs to be done in how to evaluate model-generated feedback in a on-going way that involves educational practitioners. 

% Another promising path for future research involves fine-tuning GLMs for general ATS. Using the ASAP datasets, for instance, it would be possible to employ cross validation to assess how models would perform on out-of-sample items. That is, if a GLM was fine-tuned on 9 out of 10 short answer items, how would it perform on on the tenth, held-out item. 

\bibliographystyle{plain}
\bibliography{lib}{}

\end{document}